\documentclass[a4paper]{svproc}
\usepackage{url}
\usepackage{cite}
\usepackage{graphicx}
\usepackage{xcolor, soul}
\usepackage{multirow}
\usepackage{array}
\usepackage{booktabs}
\usepackage{tikz}
\usepackage{makecell}
\usepackage{subcaption}
\usepackage{hyperref}
\usepackage{amsmath}
\usepackage{amssymb}
\usepackage{siunitx}

\usepackage{eso-pic}

\newcommand{\ISERstamp}{%
  \parbox{\textwidth}{\raggedright\bfseries\normalsize\sffamily
  Accepted for publication in the Proceedings of the\\
  19th International Symposium on Experimental Robotics (ISER)\\
  July 6--10, 2025, Santa Fe, New Mexico, USA}%
}

\newcommand{\AddFirstPageFooter}{%
  \AddToShipoutPictureFG*{%
    \AtPageLowerLeft{%
      \raisebox{2cm}{\hspace{2cm}\makebox[0pt][l]{\ISERstamp}}%
    }%
  }%
}

\begin{document}
\mainmatter              
\title{Fluidically Innervated Lattices Make Versatile and Durable Tactile Sensors}
\titlerunning{Tactile Sensing through Fluidic Innervation}  
%
\author{Annan Zhang \and Miguel Flores-Acton \and Andy Yu \and Anshul Gupta \and Maggie Yao \and Daniela Rus}
\authorrunning{Zhang, Flores-Acton, Yu, Gupta, Yao, Rus} 
%
\tocauthor{Annan Zhang, Miguel Flores-Acton, Andy Yu, Anshul Gupta, Maggie Yao, Daniela Rus}
\institute{Computer Science and Artificial Intelligence Laboratory\\ Massachusetts Institute of Technology, Cambridge, MA 02139, USA\\
\href{mailto:zhang@csail.mit.edu}{\tt zhang@csail.mit.edu}
}

\AddFirstPageFooter
\maketitle              

\begin{abstract}
Tactile sensing plays a fundamental role in enabling robots to navigate dynamic and unstructured environments, particularly in applications such as delicate object manipulation, surface exploration, and human-robot interaction. In this paper, we introduce a passive soft robotic fingertip with integrated tactile sensing, fabricated using a 3D-printed elastomer lattice with embedded air channels. This sensorization approach, termed fluidic innervation, transforms the lattice into a tactile sensor by detecting pressure changes within sealed air channels, providing a simple yet robust solution to tactile sensing in robotics. Unlike conventional methods that rely on complex materials or designs, fluidic innervation offers a simple, scalable, single-material fabrication process. We characterize the sensors' response, develop a geometric model to estimate tip displacement, and train a neural network to accurately predict contact location and contact force. Additionally, we integrate the fingertip with an admittance controller to emulate spring-like behavior, demonstrate its capability for environment exploration through tactile feedback, and validate its durability under high impact and cyclic loading conditions. This tactile sensing technique offers advantages in terms of simplicity, adaptability, and durability and opens up new opportunities for versatile robotic manipulation.
\end{abstract}

\section{Introduction}

Robots operating in unstructured environments require reliable tactile sensing to manipulate delicate objects, explore surfaces, and interact safely with humans~\cite{ kim2013soft, rus2015design}. Unlike external cameras, which suffer from occlusions and require controlled lighting, tactile sensors offer direct feedback on force, texture, and material properties~\cite{yousef2011tactile, dahiya2009tactile}. However, existing tactile sensing technologies face significant challenges in terms of durability, integration, and fabrication complexity.

Current tactile sensing approaches can be broadly categorized into piezoresistive, capacitive, optical, triboelectric, liquid metal-based, and vision-based systems. Piezoresistive and capacitive sensors, while widely used, are prone to hysteresis, drift, and environmental noise~\cite{peng2021recent, ji2020artificial}. Liquid metal-based sensors provide flexibility but face issues related to leakage and material toxicity~\cite{wang20213d, ren2020advances, qin2024emerging}. Triboelectric tactile sensors leverage triboelectric nanogenerators for self-powered sensing, but their limited output voltage restricts real-time applications~\cite{choi2023recent, lin2023self}. Vision-based tactile sensors, such as GelSight, capture high-resolution contact information but rely on deformable soft materials, like gels, which reduce longevity, and complex optical setups, which limit adaptability to different geometries~\cite{yuan2017gelsight, azulay2023allsight, quan2022hivtac, do2022densetact, do2023densetact}.

To address these limitations, we present an experimental validation of a novel tactile sensing method based on \textit{fluidic innervation}. Our approach builds on prior work on fluidically innervated structures~\cite{truby2022fluidic, zhang2024embedded, chen2024real} and introduces a 3D-printed soft robotic fingertip with embedded air channels that function as pressure-sensitive tactile sensors. Unlike conventional multi-material or vision-based tactile sensors, our fingertip is fabricated in a single step from a single material and enables high durability and seamless integration into soft robotic systems.

This work extends previous works by integrating a fluidically innervated tactile sensor with a robotic system and demonstrating its use for displacement sensing, real-time interaction, and environment exploration. We develop a model to estimate tip displacement, train a neural network for force estimation, and validate the sensors' response under various loading conditions. Additionally, we integrate the fingertip into an admittance controller to enable compliant manipulation and demonstrate its ability to explore unknown environments through tactile feedback. Finally, we evaluate its durability under high-impact and cyclic loading, highlighting its robustness for long-term robotic applications.

\section{Technical Approach}

\begin{figure}[t]
    \centering
    \includegraphics[width=1.\textwidth]{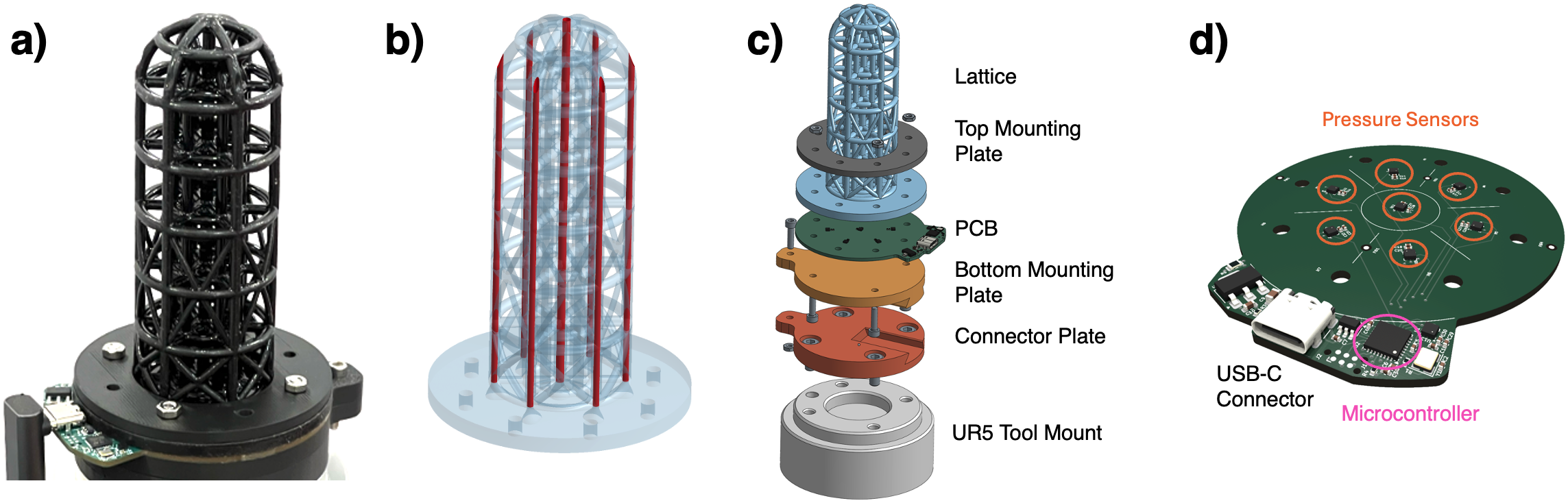}
    \caption{Sensor design. (a) Photo of the device. (b) Transparent render showing embedded air channels in red. (c) Assembly on UR5. (d) Close-up of the PCB.}
    \label{fig:technical_approach}
    \vspace{-4mm}
\end{figure}

\subsection{Lattice Design and Fabrication}

The tactile sensor is a 3D-printed elastomer lattice with embedded air channels for fluidic sensing, as shown in Fig.~\ref{fig:technical_approach}a-b. The lattice consists of two concentric circles (16 mm and 32 mm diameter), connected by six radial struts. Axial struts are placed at the center and where the radial and circular struts intersect, converging at the top into a dome. To prevent buckling, angled cross-struts were added in the lower sections, providing torsional stiffness near the base. These struts are thicker in the first section to ensure smooth bending under side loads. The axial, radial, and circular struts have a diameter of 2.7 mm, except for the reinforced central strut (3.2 mm). The air channels are 1.5 mm wide, with the central channel at 2 mm. The lattice has a vertical pitch of 16 mm and a total height of 96 mm. We designed the lattice in Onshape and fabricated it from an elastomeric polyurethane (EPU40) using a Carbon M1 resin printer. After printing, we removed uncured resin, and cured the structure in an oven. Finally, we glued the lattice onto the PCB, and sealed the air channels with epoxy.

\subsection{Readout Electronics and Assembly}

To optimize signal acquisition while minimizing tubing and environmental interference, we directly integrated LPS22HH pressure transducers onto a compact custom PCB (Fig.~\ref{fig:technical_approach}d), positioned close to the air channels. This design reduces channel volume, improves temperature stability, and significantly reduces system bulk, compared to the use of rubber tubing and off-the-shelf readout electronics presented in previous work~\cite{truby2022fluidic,zhang2024embedded,chen2024real}. The pressure transducers interface with an STM32 microcontroller via SPI, which transmits pressure and temperature data at 200 Hz over USB. The entire assembly, including the lattice, PCB, and mounting plates, is secured with screws and attaches to the UR5 robot arm via a dovetail connector for rapid interchangeability (Fig.~\ref{fig:technical_approach}c).

\section{Experimental Evaluation}

\subsection{Characterization}
\label{sec:characterization}

\begin{figure}
\vspace{-4mm}
  \centering
  \includegraphics[width = 1.\textwidth]{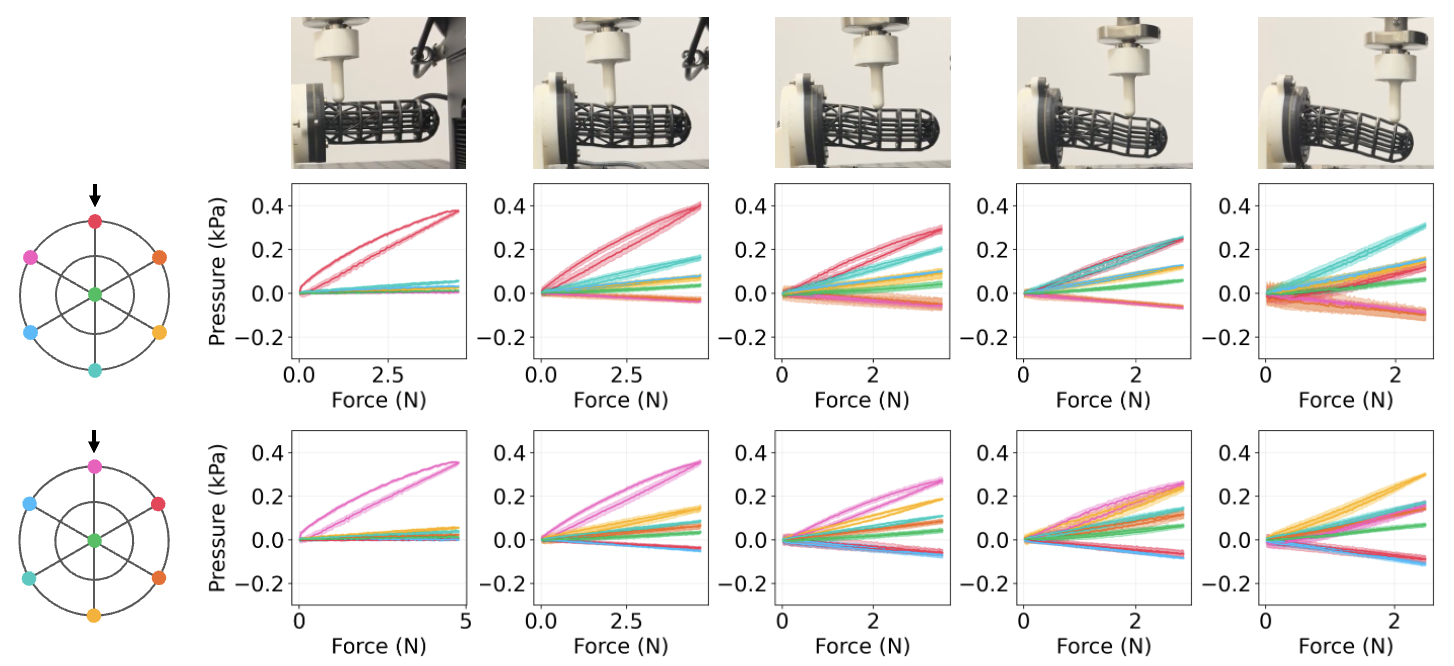}
  \caption{Characterization of sensor responses on the mechanical testing machine, showing force application at five positions along the axial direction and six orientations (two displayed here). Each plot shows the mean sensor response across five trials, with shaded regions representing ±1 standard deviation. Responses are color-coded by sensor.}
  \label{fig:char}
\end{figure}

The sensing mechanism is governed by the ideal gas law, $pV = nRT$. When confined to a closed volume, changes in pressure within the air channels can be measured as the lattice deforms. For a discussion of leakage and temperature compensation methods, refer to previous work~\cite{truby2022fluidic,zhang2024embedded}.

To characterize the tactile sensor's response, we applied normal forces using a universal testing machine (Instron). A 3D-printed PLA tip was used to apply compression at five discrete axial positions along the lattice, as well as six different angular orientations corresponding to the locations of the air channels' struts. The resulting sensor responses as a function of the applied compression force are shown in Fig.~\ref{fig:char}. Each column corresponds to one of the five axial positions, with the top row images showing the Instron test setup and the five positions. Each row of plots represents one of the six orientations tested (two shown). We report the mean sensor response across five trials for all seven sensors, with shaded areas denoting one standard deviation. The sensor responses are color-coded to match the air channel colors in the diagrams on the left.
We make two key observations:

\textbf{Position-dependent sensor response}: When force is applied near the base (first column), the top channel shows a strong response, while other channels remain mostly unresponsive. Conversely, when force is applied near the tip (last column), the strongest response comes from the diametrically opposite channel (i.e., the bottom), as the lattice bends downward. The top channel experiences a superposition of direct compression (positive pressure) and tension from bending (negative pressure), causing a weaker net response. For intermediate positions, we observe a gradual transition between these extremes, with the top channel's response weakening and the bottom channel's response strengthening as force moves from the base to the tip. This behavior is consistent with previous observations in fluidically innervated grippers, where a superposition of direct compression and bending deformations was identified~\cite{zhang2024embedded}.

\textbf{Orientation-dependent symmetry}: Due to the 60-degree rotational symmetry of the lattice and air channel structure, we expect qualitatively similar sensor responses for different orientations, with only the specific channels activated varying. This is confirmed by the similarity in sensor responses between the first and second rows, where the lattice was rotated by 60 degrees clockwise.

\subsection{3D Tip Displacement Estimation}
\label{sec:geometric}

\begin{figure}
  \centering
  \includegraphics[width = 1.\textwidth]{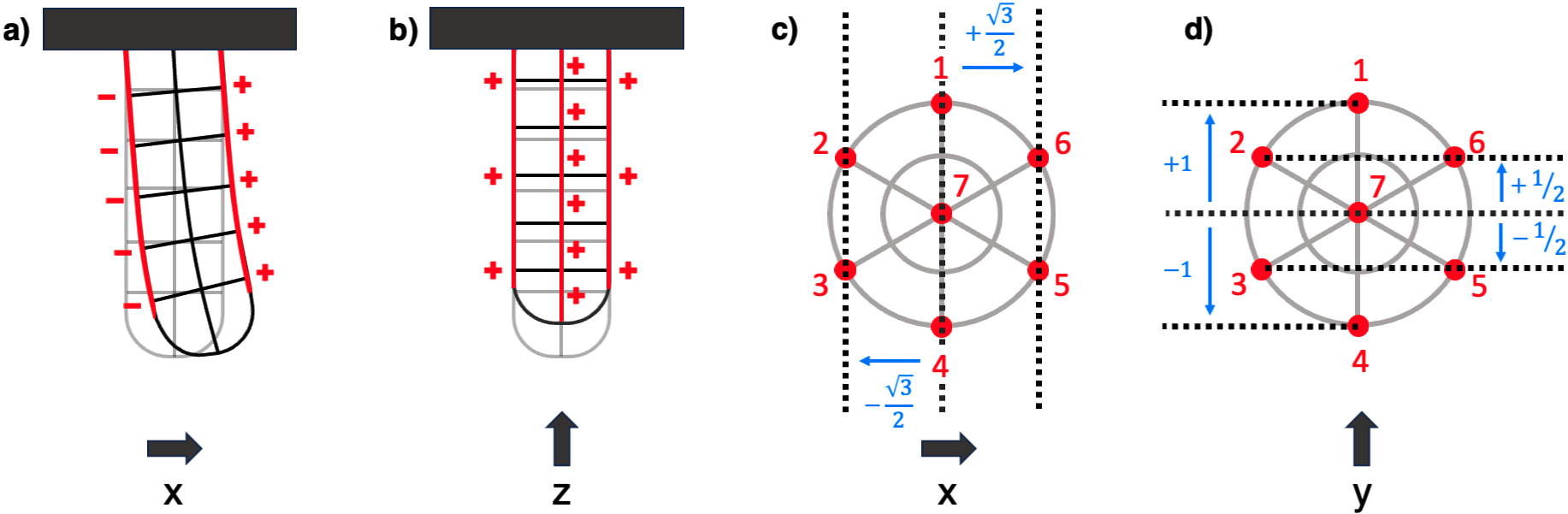}
  \caption{Geometric modeling approach for estimating tip displacements.}
  \label{fig:geometry}
  \vspace{-4mm}
\end{figure}

\begin{figure}[b]
\vspace{-4mm}
    \centering
    \includegraphics[width=1.\textwidth]{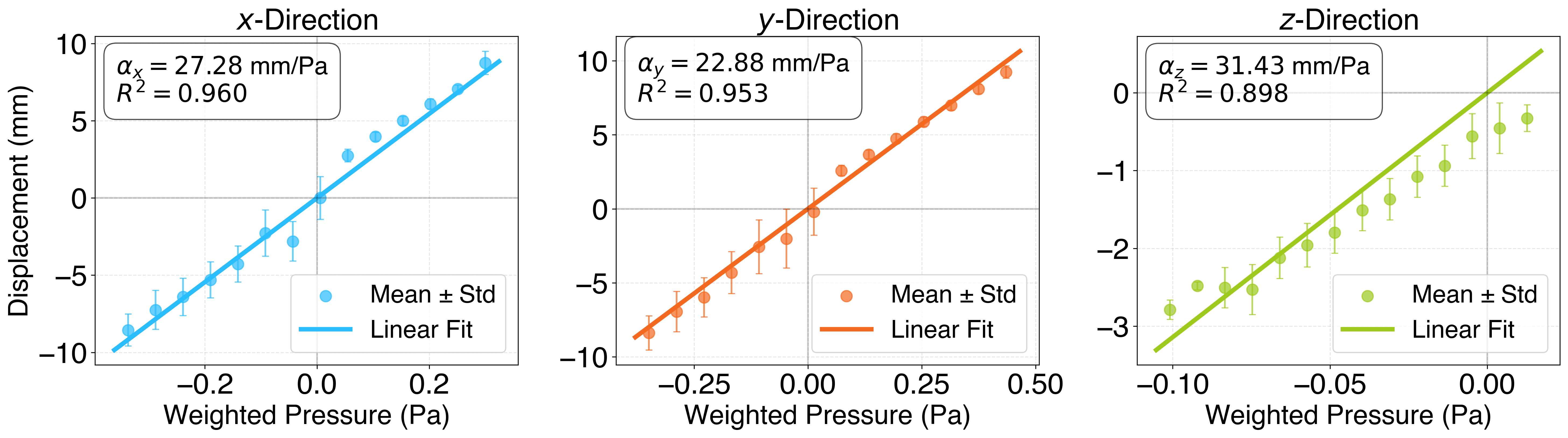}
    \caption{Experimental validation of the linear relationship between weighted pressure readings and tip displacements, proposed in Eqns.~\eqref{eq:delta_x}--\eqref{eq:delta_z}.}
    \label{fig:model_valid}
\end{figure}

For external forces acting at the fingertip, we develop a geometric approach to estimate the 3D displacement of the fingertip. 
An external force applied in the positive $x$-direction induces bending that moves the tip in the positive $x$-direction, leading to positive pressure readings on the compression side and negative readings on the tension side (see Fig.~\ref{fig:geometry}a). This behavior aligns with observations in previous work~\cite{truby2022fluidic}. We evaluate each pressure channel: the more its pressure value matches the expected pattern under a positive $x$-force (e.g., positive pressures for channels 5 and 6, negative pressures for channels 2 and 3 as shown in Fig.~\ref{fig:geometry}c), the more confident we are of a force in the positive $x$-direction. The magnitude of the force correlates with the magnitude of the measured pressure change~\cite{zhang2024embedded}.
The extent to which each channel is compressed or stretched depends on its normal distance from the neutral plane (depicted as the vertical dotted line in the center of Fig.~\ref{fig:geometry}c). Therefore, we weight each sensor reading by its normal distance from the neutral plane in the cross-section. A similar argument applies for the $y$-direction, but the weights differ due to the 60-degree separation of the channels (see Fig.~\ref{fig:geometry}d).
When the tip moves in the $z$-direction, the direct compression of the center strut results in a positive reading in the channel 7 (see Fig.~\ref{fig:geometry}b). To ensure that effects on the displacement estimate cancel out if all channel pressures increase by the same constant amount (e.g., due to ambient temperature changes in a closed volume), the weights are designed to sum to zero. This approach leads to the following equations for estimating the tip displacements:
\begin{align}
\delta_x &= \alpha_x \left( \frac{\sqrt{3}}{2} (p_5 + p_6 - p_2 - p_3) \right) \label{eq:delta_x} \\
\delta_y &= \alpha_y \left( p_1 - p_4 + \frac{1}{2} (p_2 - p_3 - p_5 + p_6) \right) \label{eq:delta_y} \\
\delta_z &= \alpha_z \left( p_7 - \frac{1}{6} (p_1 + p_2 + p_3 + p_4 + p_5 + p_6) \right) \label{eq:delta_z}
\end{align}
where $\delta_i$ are the estimated displacements from the pressure readings, $\alpha_i$ are scaling parameters to be identified, and $p_1, p_2, \ldots, p_7$ are the deformation-induced pressure changes.
Experimental validation using controlled displacements with an Instron machine shows a strong linear relationship between displacements and weighted pressures (parentheses in Eqns.~\eqref{eq:delta_x}--\eqref{eq:delta_z}), up to ±10\,mm in the $x$- and $y$-directions and 3\,mm in the negative $z$-direction (see Fig.~\ref{fig:model_valid}).

\subsection{Contact Location and Force Detection}

\vspace{-4mm}
\begin{figure}
  \centering
  \includegraphics[width = .5\textwidth]{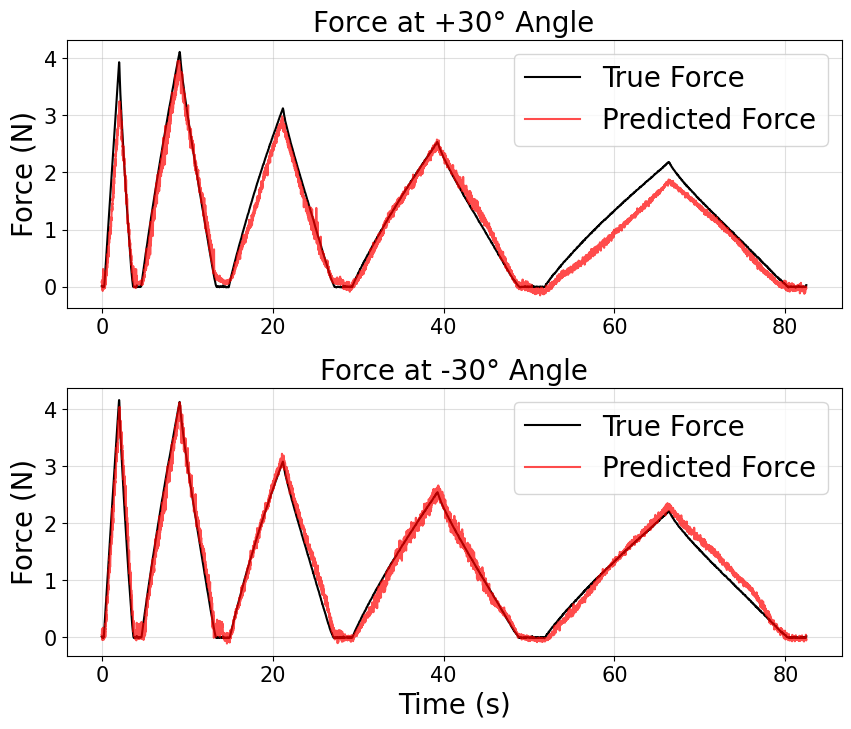}
  \caption{Predicted force vs. ground truth for 5 positions at 30$^\circ$ and -30$^\circ$, showing strong prediction accuracy for both positions.}
  \label{fig:force}
  \vspace{-4mm}
\end{figure}

Detecting forces acting along the cylindrical part of the device is more complex than detecting those at the tip. Deformations in this region compress individual channels, leading to highly non-linear and time-dependent effects. As shown in Section~\ref{sec:characterization}, even applying normal forces with the Instron results in distinct sensor signals. The material's large deformations and non-linear constitutive behavior further complicate modeling. However, based on the characterization data, we observe a clear qualitative difference between contacts at the top and bottom, making it feasible to estimate the axial force location. The radial force location can be deduced from which channels are activated, as they are positioned differently around the circumference. Furthermore, higher forces generally produce stronger sensor signals, allowing us to estimate force magnitude.

For this highly complex task of predicting contact location and forces, we use a machine learning-based approach. We train a neural network (NN) trained on data from our characterization experiments, where we applied forces at 5 axial positions and 6 radial angles, repeated over 5 trials. The NN takes in the 7 sensor values at each timestep and predicts the current contact location (as one of 6 radial angles and one of 5 axial positions) and the force magnitude.
Following the approach in~\cite{zhang2023machine}, we collect an additional trial as a test set to validate the model. We use a fully-connected NN with 2 hidden layers, each with 128 neurons. The model, containing 37K parameters, trains in under 3 minutes on a desktop PC with an RTX 3090 GPU. On the test set, the NN achieves a classification accuracy of 95\% for the axial position, 99\% for the radial angle, and a mean absolute error (MAE) of 0.16 N for force predictions. Fig.~\ref{fig:force} shows predicted force vs. ground truth for 5 positions at two angles: one at 30$^\circ$ (with a force MAE of 0.17 N), and one at -30$^\circ$ (with a force MAE of 0.12 N).
The NN's forward pass is extremely fast (0.29 $\pm$ 0.04 ms), making it suitable for real-time applications. This processing speed allows us to fully utilize the 200 Hz data rate of our sensor, a common issue with many vision-based solutions.

While our setup demonstrates promising results, additional work is required to adapt this approach for real-world manipulation tasks, where contact patches are more complex (e.g., larger areas instead of point contacts), and forces act in arbitrary directions. This experiment shows that even with a simple NN and no historical data, we can extract rich and accurate information about contact location and applied force from only 7 sensor channels.

\subsection{Admittance Control with Tunable Stiffness}
\label{sec:admittance}

\vspace{-4mm}
\begin{figure}
  \centering
  \includegraphics[width = 1.\textwidth]{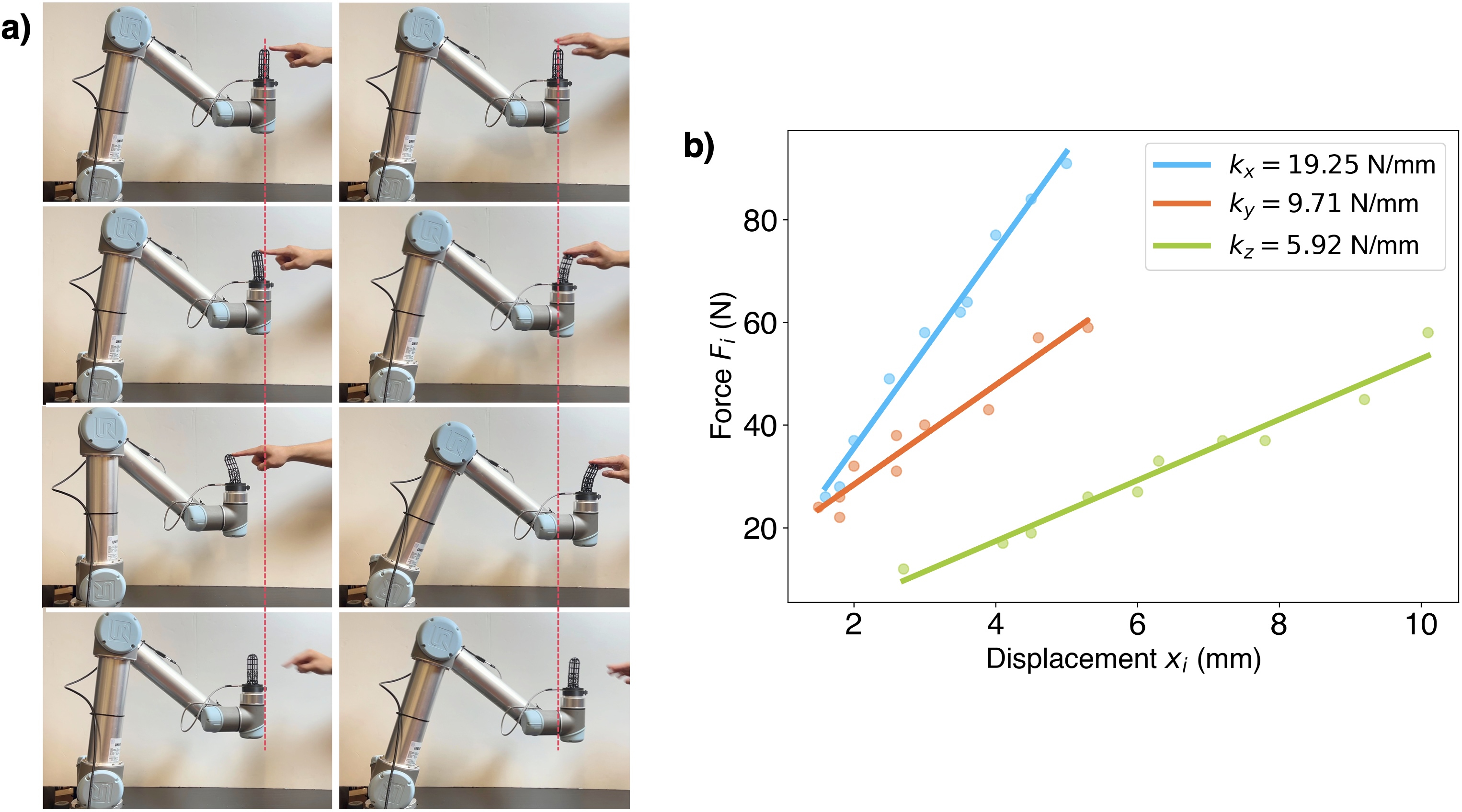}
  \caption{Admittance controller demonstrating spring-like behavior of the robot tip. (a) Sequences of images from the supplementary video. As a horizontal force is applied and then removed, the robot moves back towards its original position (marked by dashed red line). (b) Effective stiffness measurements in each direction for gains chosen in Eqn.~\eqref{eq:beta_values}, with linear least squares fit.}
  \label{fig:admittance}
  \vspace{-4mm}
\end{figure}

To demonstrate the fingertip's capabilities, we implement an admittance control scheme that enables the robot arm's end-effector to behave like a linear spring. We require the force to displace the tip to increase linearly with the displacement, with a spring stiffness that can be individually tuned along each axis ($k_x$, $k_y$, $k_z$).
Since our robot (UR5) is position-controlled, we respond to external forces applied at the fingertip by reading the pressure signals, estimating the tip displacements using the geometric approach from Section~\ref{sec:geometric}, and computing position commands for the UR5. The control law is defined as:
\begin{equation}
u_i = \beta_i \delta_i, \quad \text{for } i \in \{ x, y, z \}
\label{eq:control_law}
\end{equation}
where $u_i$ is the commanded position of the UR5 end-effector in the $i$-th direction, $\delta_i$ is the estimated tip displacement from Eqns.~\eqref{eq:delta_x}--\eqref{eq:delta_z}, and $\beta_i$ is the control gain for the $i$-th direction.
To achieve a natural interaction feel, we select:
\begin{equation}
\alpha_x\beta_x = \alpha_y\beta_y = \frac{1}{15}, \quad \alpha_z\beta_z = \frac{1}{7.5}
\label{eq:beta_values}
\end{equation}
With these gains, the robot responds to fingertip displacements in a way that simulates a virtual spring, providing tactile feedback to the user.
We experimentally determine the effective stiffnesses $k_x$, $k_y$, and $k_z$ corresponding to these gain settings. For each direction, we apply 10 different forces $F_i$ to the tip using a handheld force gauge and record the corresponding total displacements $x_i = \delta_i + u_i$. We then plot the force vs. displacement data and perform a linear least squares fit to determine the effective stiffness as $k_i = \frac{F_i}{x_i}$.
The stiffness measurements are displayed in Fig.~\ref{fig:admittance}b, along with the linear fits. Fig.~\ref{fig:admittance}a illustrates the spring-like behavior of the robot tip under the admittance controller. When the user applies a horizontal force, the robot moves back, and upon releasing the force, the robot returns to its original position. We show clips of different interactions in the supplementary video.

\subsection{Tactile Exploration and Mapping}
\label{sec:exploration}

\vspace{-6mm}
\begin{figure}[h!]
  \centering
  \includegraphics[width = .9\textwidth]{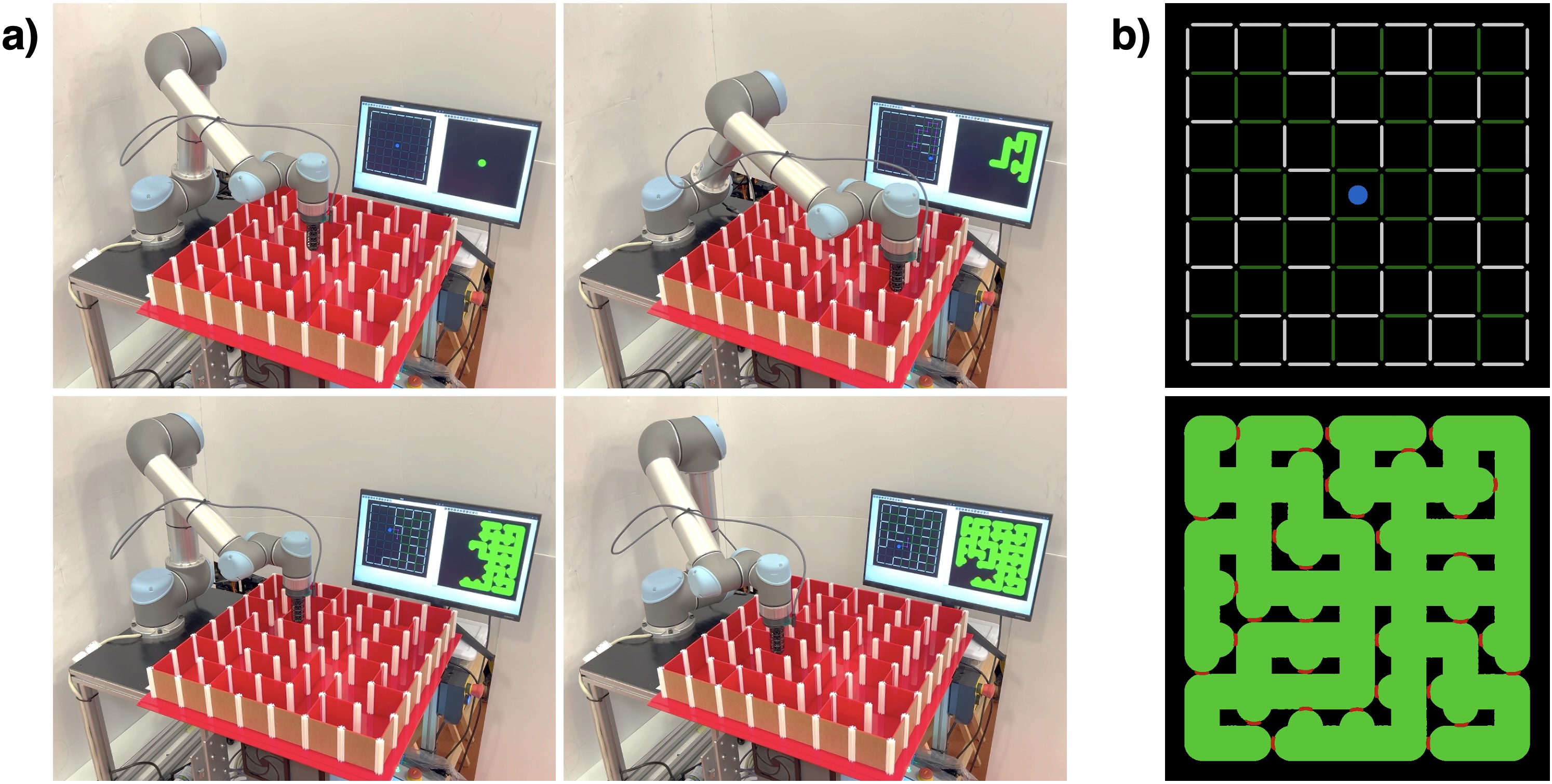}
  \caption{Tactile exploration of a maze. (a) Robot stages while exploring and the evolving knowledge of the maze displayed on a monitor. (b) Final exploration result, showing inferred walls (white) and open spaces (green) on the top, and lattice traversed regions (green) and contact points (red) on the bottom.}
  \label{fig:maze}
\end{figure}

We demonstrate our tactile sensor's ability to explore and map unknown environments in a 2D maze task. To detect obstructions and open spaces, the environment is divided into 1 mm $\times$ 1 mm cells for an effective balance between resolution and efficiency. Fig.~\ref{fig:maze} bottom row shows a heuristic cell map (right) and a simplified maze abstraction (left). This approach can extend to more complex or 3D environments, but here we focus on demonstrating effective tactile exploration with our fingertip, using the geometric model from Section~\ref{sec:geometric}. The contact surface is determined by the lattice radius, tip displacement, and thickness parameters, with $\delta_x$ and $\delta_y$ (from Eqns.~\eqref{eq:delta_x}--\eqref{eq:delta_y}) estimating contact location. The contact angle is computed using $\arctan_2(\delta_x, \delta_y)$, and an arc with radius and span defines the contact surface.

For maze exploration, the robot moves in a constrained $x$- or $y$-direction, advancing as long as the tip displacement stays below a threshold. If the sensor detects an obstruction, the robot marks the edge as a wall and returns to the previous cell. Otherwise, the edge is marked clear, and the new cell is added to the stack. The maze is explored using a Depth First Search (DFS) algorithm, with the robot checking neighboring cells based on three criteria: the cell is within the maze boundaries, not in the stack, and connected by an undefined wall. Boundary edges are treated as walls. The robot recursively explores unexplored cells, backtracking when no valid option remains. We repeat the experiment with various maze configurations, and the robot consistently identifies all walls correctly (see supplementary video).

\subsection{Durability Against Impact and Fatigue}
\label{sec:durability}

\vspace{-4mm}
\begin{figure}
  \centering
  \includegraphics[width = 1.\textwidth]{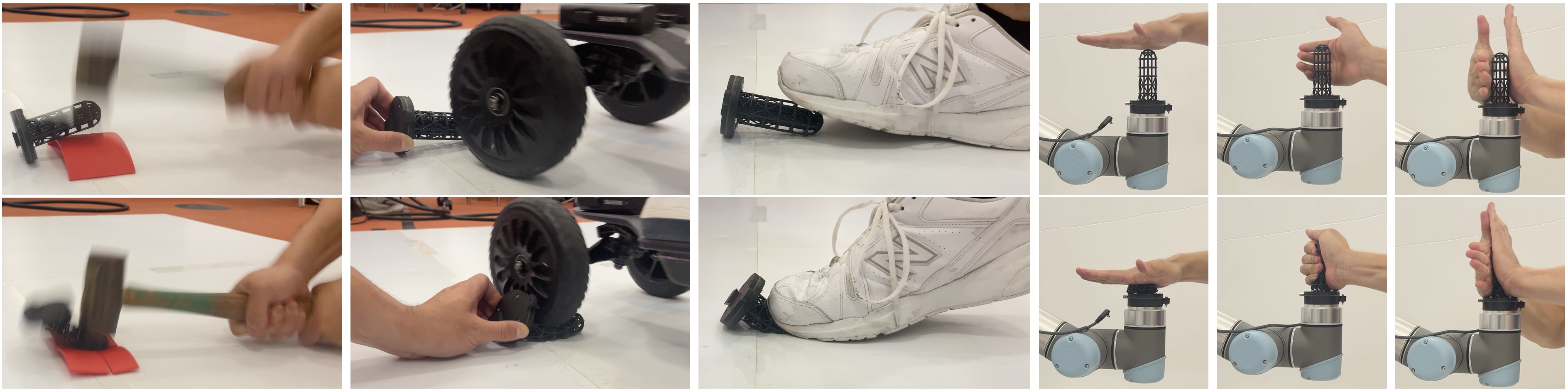}
  \caption{The lattice subjected to high-impact loading and extreme deformations, including squeezing and stepping, as well as being hammered and run over by an electric skateboard.}
  \label{fig:impact}
  \vspace{-4mm}
\end{figure}

We demonstrate the durability of the lattice and the entire device by subjecting it to extreme impact and deformation tests. Shown in Fig.~\ref{fig:impact}, we compressed the lattice top-down, side-to-side, and radially while mounted. Additionally, we subjected the device to extreme loading, with an 80 kg individual stepping on it and a person riding over it on an electric skateboard (weight 70 kg). To test resistance to high peak forces, we struck the device with a hammer, generating enough impact to crack a 3 mm thick PLA sheet beneath it. Uncut video clips in the supplementary material show the full sequence, where we ran the admittance control demo from section~\ref{sec:admittance}, unplugged and dismounted the device, subjected it to these extreme tests, and remounted it on the robot arm to verify continued function.
Over the course of filming, the same device endured repeated hammer blows, stepping, and skateboard tests. After further intentional stress, mechanical damage rendered it unmountable, but the tactile sensing still worked and could be used to control the UR5. The device ultimately failed when additional hammer strikes severed PCB traces.

\begin{figure}
  \centering
  \includegraphics[width = .5\textwidth]{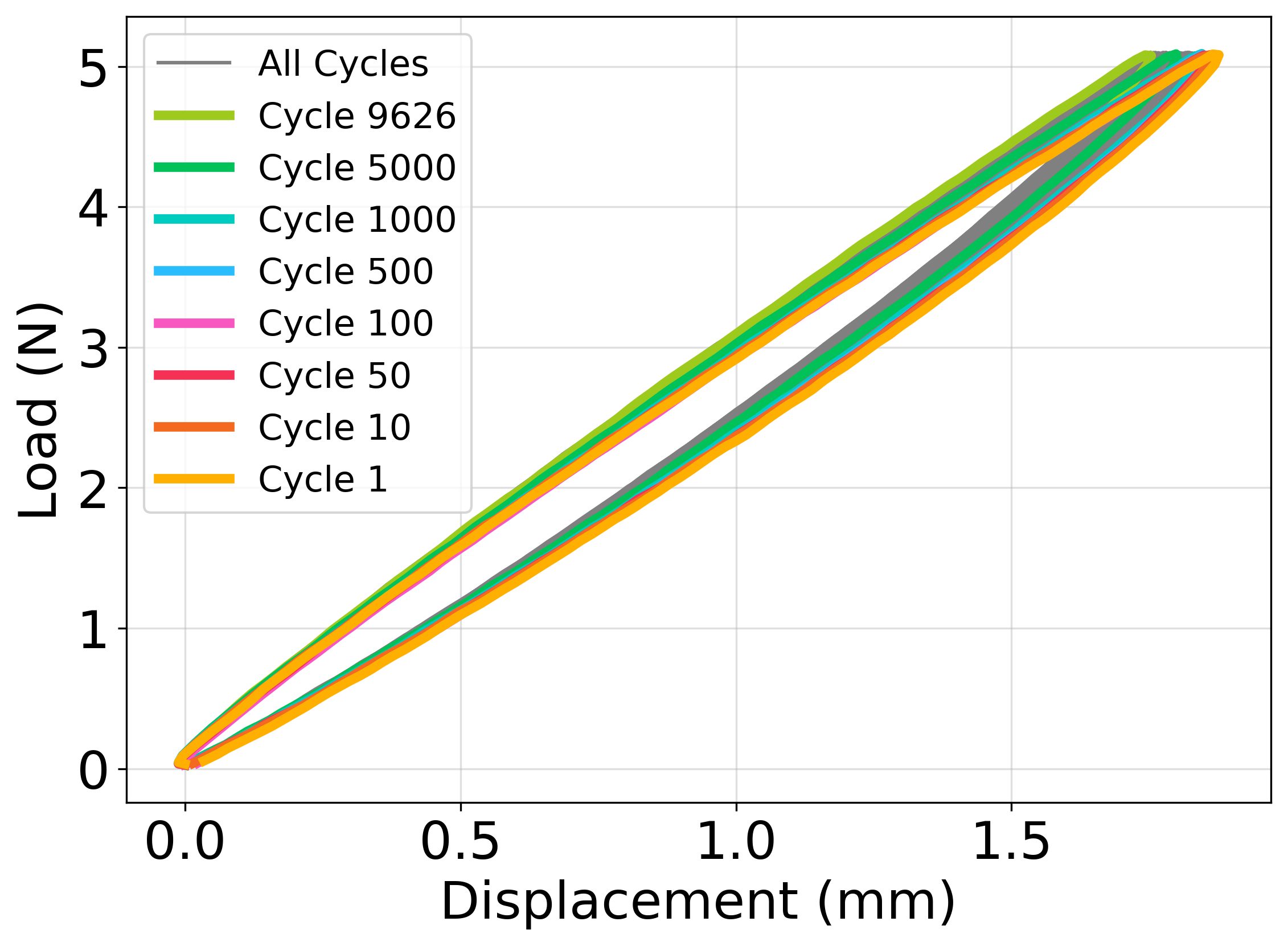}
  \caption{Force-displacement cycles during fatigue testing.}
  \label{fig:fatigue}
  \vspace{-4mm}
\end{figure}

To evaluate the device's durability under fatigue, we subjected it to cyclic loading using an Instron machine. The Instron was set to 1 mm/s for 10,000 cycles, applying forces from 0.05 to 5.05 Newtons to maintain contact (with the setup as in Fig.~\ref{fig:char} top left plot). The displacement gauge was reset before each loading cycle. The first five cycles were discarded due to stress softening of the elastomeric polyurethane. After filtering out cycles with NaN values, 9,626 cycles were retained for analysis.
The resulting force-displacement plot over all cycles is shown in Fig.~\ref{fig:fatigue}. All cycles are plotted in grey, with the 1st, 50th, 100th, 500th, 1,000th, 5,000th, and final 9,626th cycles highlighted in color. Remarkably, the cycles overlap almost entirely, even after nearly 10,000 repetitions, suggesting that the lattice is highly durable against fatigue. However, further work is needed to assess how the mechanical properties and sensor responses evolve over prolonged cycling, and how many cycles the device can ultimately withstand.

\section{Conclusion}

The introduction of fluidic innervation in a passive soft robotic fingertip provides a simple yet effective solution to tactile sensing, offering enhanced adaptability and durability. This approach leverages a scalable, single-material fabrication process that detects pressure changes within air channels, eliminating the complexity of conventional methods. By integrating the fingertip with an admittance controller and validating its performance under various conditions, we demonstrate its potential for a wide range of robotic applications, from force sensing to environmental exploration. This work opens up new opportunities for more robust and versatile robotic manipulation in unstructured and dynamic environments.

\paragraph{\textbf{Acknowledgments.}} This work was supported by the Singapore-MIT Alliance for Research and Technology (SMART) Mens, Manus, and Machina program and the Gwangju Institute of Science and Technology.

\paragraph{\textbf{Supplementary Video.}} A video demonstrating the tactile sensor's capabilities, including admittance control, maze exploration, and durability tests, is available at: \url{https://youtu.be/sOCviB8Zi3Q}

\bibliographystyle{splncs03_unsrt}
\bibliography{library}

\end{document}